\documentclass[OJMO,NoBabel,manuscript]{cedram}
\usepackage{pgfplots}
\usepackage{algorithm}
\usepackage{algorithmicx}
\usepackage[noend]{algpseudocode}
\usepackage{graphicx}

\usepackage{xcolor}
\hypersetup{colorlinks}

\usepackage{enumerate}
\usepackage{tikz-cd}

\newcounter{theorem}

 \newtheorem{definition}[theorem]{Definition}

 \newtheorem{assumption}[theorem]{Assumption}
\newcommand{\beq}{\begin{equation}}
\newcommand{\eeq}{\end{equation}}

\definecolor{amethyst}{rgb}{1, 0, 1}
\definecolor{blue-violet}{rgb}{0.54, 0.17, 0.89}
\definecolor{brightturquoise}{rgb}{0.03, 0.91, 0.87}

\newcommand{\ip}[1]{\left\langle #1 \right\rangle}
\newcommand{\T}{^{\mathsf{T}}}
\newcommand{\lt}{\left}
\newcommand{\rt}{\right}
\newcommand{\one}{\mathbf{1}}
\newcommand{\mat}[1]{\begin{bmatrix} #1 \end{bmatrix}}
\newcommand{\cD}{\mathcal{D}}
\newcommand{\R}{\mathbb{R}}
\newcommand*\Let[2]{\State #1 $\gets$ #2}
\newcommand{\prox}{\text{prox}}
\title{Estimating Shape Parameters of Piecewise Linear-Quadratic Problems}

\author{\firstname{Peng}  \lastname{Zheng}}
\address{Department of Health Metrics Sciences, University of Washington\\
  Seattle, WA}
\email{zhengp@uw.edu}
  
\author{\firstname{Karthikeyan} \middlename{N.}  \lastname{Ramamurthy}}
\address{IBM T.J. Watson Research Center\\ Yorktown Heights, NY}
\email{knatesa@us.ibm.com}

\author{\firstname{Aleksandr} \middlename{Y.} \lastname{Aravkin}}
\address{Department of Applied Mathematics, University of Washington \\ Seattle, WA}
\email{saravkin@uw.edu}

 \thanks{The authors acknowledge ***.}

\date{\today}

\begin{document}

\maketitle



%\twocolumn[
%
%\aistatstitle{Unified Optimization for Shape Parameter Estimation}
%
%\aistatsauthor{ Peng Zheng \And Aleksandr Aravkin \And  Karthikeyan Ramamurthy }
%%
%\aistatsaddress{ Applied Mathematics \\ University of Washington \And Applied Mathematics \\ University of Washington \And IBM TJ Watson Research Center } 
%
%]

\begin{abstract}
Piecewise Linear-Quadratic (PLQ) penalties are widely used to develop models in statistical inference, signal processing, and machine learning. 
Common examples of PLQ penalties include least squares, Huber, Vapnik, 1-norm, and their asymmetric generalizations. 
Properties of these estimators depend on the choice of penalty and its shape parameters, 
such as degree of asymmetry for the quantile loss, and transition point between linear and quadratic pieces for the Huber function. 

In this paper, we develop a statistical framework that can help the modeler to automatically tune shape parameters once the shape of the penalty 
has been chosen. The choice of the parameter is informed by the basic notion that each QS penalty should correspond to a true statistical density. 
The normalization constant inherent in this requirement helps to inform the optimization over shape parameters, giving a joint optimization problem 
over these as well as primary parameters of interest. A second contribution is to consider optimization methods for these joint problems. 
We show that basic first-order methods can be immediately brought to bear, and design specialized extensions of interior point (IP) methods 
for PLQ problems that can quickly and efficiently solve the joint problem. 
Synthetic problems and larger-scale practical examples illustrate the potential of the approach. 

%In this paper, we optimize over parametrized spaces of penalties to simultaneously learn the model and the misfit penalty.
%We discuss theoretical properties of such problems, and {develop} algorithms for their solution. The ideas are illustrated using synthetic examples of optimizing over the space of piecewise linear-quadratic (QS) penalties, and the approach is applied to develop a self-tuning {robust PCA} formulation for background separation.

\end{abstract}

%======================================================================
\section{Introduction}

Piecewise Linear-Quadratic (PLQ) functions and their superclass, Quadratic Support functions~\cite{JMLR:v14:aravkin13a} form a rich class of penalties that are used 
for a range of applications. The choice of penalty plays a key role in the properties of the final estimate. 
Using the Huber penalty (Figure~\ref{fig:Qhub}(a)) rather than a quadratic to measure 
data misfit is a typical approach to make an estimate robust to outliers~\cite{huber2004robust}. The 1-norm applied
directly to parameters makes the final solution sparse, and is used in Lasso regression~\cite{tibshirani1996regression} and compressed sensing~\cite{donoho2006compressed}. 
The asymmetric quantile loss (see Figure~\ref{fig:Qhub}(b)) is used for many regression applications~\cite{koenker2001quantile}. 

The correspondence of the {\it shape} of a PLQ function to its role in estimation is well-understood. 
For example, it is the linear tails of the Huber penalty that limit the effects of large residuals on the final estimate 
when used as a misfit, and it is the nonsmooth behavior at the origin of the 1-norm that promotes sparse solutions
when used as a regularizer. Likewise, the asymmetry of the quantile loss that make it useful for financial applications, 
where loss and gain are treated asymmetrically.

\begin{figure}
\begin{tabular}{p{3.6cm}p{3.6cm}p{3.6cm}}
%%%%%%%%%%%%%%%%%%%%%
% Densities
%%%%%%%%%%%%%%%%%%%%%
\centering
   \begin{tikzpicture}
  \begin{axis}[
    thick,
    height=2cm,
    xmin=-2,xmax=2,ymin=0,ymax=1,
    no markers,
    samples=50,
    axis lines*=left, 
    axis lines*=middle, 
    scale only axis,
    xtick={-1,1},
    xticklabels={$-\kappa$, $\kappa$},
    ytick={0},
    ] 
\addplot[red,domain=-2:-1,densely dashed]{-x-.5};
\addplot[blue, domain=-1:+1]{.5*x^2};
\addplot[red,domain=+1:+2,densely dashed]{x-.5};
\addplot[blue,mark=*,only marks] coordinates {(-1,.5) (1,.5)};
  \end{axis}
\end{tikzpicture}
& \begin{tikzpicture}
  \begin{axis}[
    thick,
    height=2cm,
    xmin=-2,xmax=2,ymin=0,ymax=1,
    no markers,
    samples=50,
    axis lines*=left, 
    axis lines*=middle, 
    scale only axis,
    xtick={-1,1},
   xticklabels={-1,1},
    ytick={0},
    ] 
\addplot[red,domain=-2:0,densely dashed]{-.3*x};
\addplot[red,domain=0:+2,densely dashed]{.7*x};
\draw[color=black] (20,30) node {$-\tau$};
\draw[color=black] (330,30) node {$1-\tau$};
  \end{axis}
\end{tikzpicture} 
&  \begin{tikzpicture}
  \begin{axis}[
    thick,
    height=2cm,
    xmin=-1.5,xmax=1.5,ymin=0,ymax=4,
    no markers,
    samples=100,
    axis lines*=left, 
    axis lines*=middle, 
    scale only axis,
    xtick={-0.45,1.05},
     xticklabels={\small$-\tau\kappa$,\small$(1-\tau)\kappa$},
    ytick={0},
    ] 
\addplot[red,domain=-1.5:-3*0.3*0.5,densely dashed]{3*0.3*abs(x)*2 - 0.5*3^2*0.3^2};
\addplot[blue,domain=-3*0.3*0.5:3*0.7*0.5]{2*x^2};
\addplot[red,domain=3*0.7*0.5:1.5,densely dashed]{3*0.7*abs(x)*2 - 0.5*0.7^2*3^2};
\addplot[blue,mark=*,only marks] coordinates {(-0.45,0.405) (1.05,2.205)};
\end{axis}
\end{tikzpicture} \\
\centering\footnotesize
(a) Huber ($\kappa$)  &\footnotesize (b) quantile ($\tau$)  & \footnotesize quantile Huber ($\tau$, $\kappa$).
\end{tabular}
\caption{\small\label{fig:Qhub} Huber and quantile families parametrized by $\kappa$ and $\tau$, with the quantile Huber parameterized by both.}
%\vspace{-.5in}
%\caption{
%Densities $\mathbf{p}(\textcolor{blue}{v})$, penalties $-\ln\mathbf{p}(\textcolor{blue}{v})$ ,
 %and influence fns $-\frac{d}{d \textcolor{blue}{v} }\ln\mathbf{p}(\textcolor{blue}{v})$. }
\end{figure}

Here, we consider the choice of parameters that fully specify this shape. At what value should the Huber transition between 
quadratic and linear?  How asymmetric do we need the quantile loss to be? These questions are easily answered
once we have residuals and estimates in hand, but they seem as difficult {\it a priori} as any other parameter selection problem.

Unknown parameters in are usually estimated using cross-validation or grid search. Standard 
methods require multiple solutions of any given learning problem, where a held-out dataset is used 
to evaluate each configuration~\cite{arlot2010survey}. 
More recently, Bayesian optimization \cite{snoek,hutter,bergstra,fastBO} and random search \cite{randSearch,randSA} have 
come to the forefront as two popular techniques used to obtain meta-parameters in a very wide range of contexts. 

While these techniques can be used to for the problems we consider, as well as a much more general problem class, 
%to discover the best shape parameters for a given error model. 
their implementation is more expensive than solving a single problem instance --- cross-validation, random search and Bayesian optimization require many instance evaluations. 

%In contrast, our approach requires solving a {\it single} extended problem to simultaneously fit the $x$ and $\theta$.

Here we take a different tack, and build on the statistical perspective developed in~\cite{JMLR:v14:aravkin13a} to develop 
a simple modification of the PLQ estimation problem to infer shape parameters simultaneously with variables of interest. 
The most relevant works related to this paper focus on the relation between the quantile penalty and the asymmetric Laplace 
distribution(ALD)~\cite{yu2001bayesian, Tu2017, bera2016asymmetric}.
In particular, \cite{bera2016asymmetric} jointly estimate the model and the shape parameters for quantile penalty,
and~\cite{Tu2017} infer the joint posterior distribution of these parameters. 

To explain the general idea, we look at two simple examples. 

\subsection{Classic variance estimation in linear regression.}
As a warm-up, we consider the classic problem of estimating regression variables jointly with noise variance parameter in a linear Gaussian model. 
Take the basic statistical model 
\begin{equation}
\label{eq:LSstat}
y = Ax + \epsilon, \quad \epsilon \sim N(0, \sigma^2 I), \quad i = 1, \dots, n,
\end{equation}
we can find the maximum likelihood estimator for $(x, \sigma^2)$ by maximizing 
\[
p(x, \sigma^2 |y) \propto p(y|x,\sigma2) = \prod_{i=1}^n \frac{1}{\sqrt{2\pi\sigma^2}}\exp((y_i - \langle a_i, x\rangle)^2/2\sigma^2) 
\]
or equivalently by minimizing the negative log of this expression: 
\begin{equation}
\label{eq:LSvar}
\min_{x, \sigma^2} \frac{1}{2\sigma^2}\|y - Ax\|^2 + n\log(\sigma^2). 
\end{equation}
In this simple case, the problem separates, and regardless of $\sigma^2$ we have
\[
\hat x = (A^TA)^{-1}A^Tb.
\]
Minimizing~\eqref{eq:LSvar} with respect to $\sigma^2$, we also get 
\[
\hat\sigma^2 = \|y - A\hat x\|^2/n,
\]
the average of empirical residuals squared.
The main point to here is that the likelihood arising from the statistical model~\eqref{eq:LSstat} enables us to estimate $\sigma^2$. 
While the two variables separate in this case, in general the picture is more interesting, as we see in the next example. 

\subsection{Simultaneous quantile estimation and  regression.}
\label{ex:quantileReg}

\begin{figure}
\includegraphics[width=6.8cm]{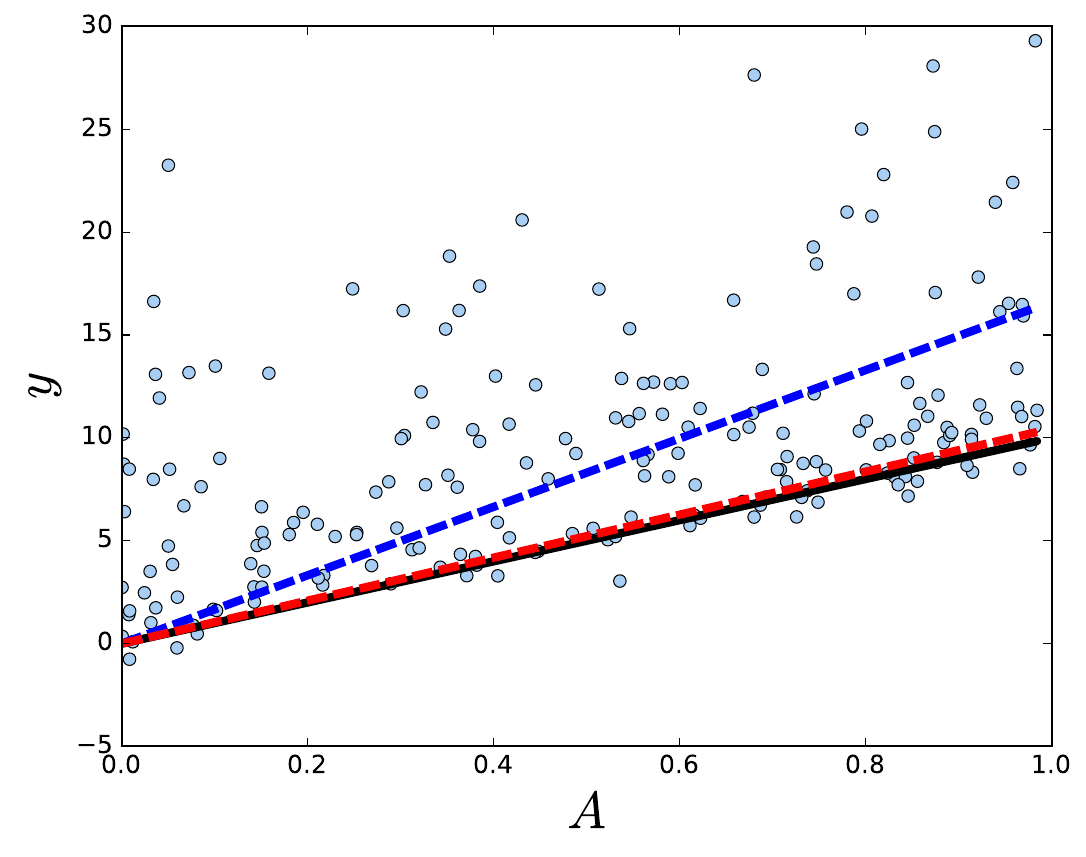}\qquad
\centering
   \begin{tikzpicture}
  \begin{axis}[
    thick,
    height=5cm,
    xmin=0,xmax=1,ymin=0,ymax=4,
    no markers,
    samples=200,
    axis lines*=left, 
    axis lines*=middle, 
    scale only axis,
    xtick={0,1},
   xticklabels={$0$, $1$},
    ytick={0},
    ] 
\addplot[red,domain=0:1,densely dashed]{ln(1/x+1/(1-x))};
%\addplot[blue, domain=-1:+1]{.5*x^2};
%\addplot[red,domain=+1:+2,densely dashed]{x-.5};
%\addplot[blue,mark=*,only marks] coordinates {(-1,.5) (1,.5)};
  \end{axis}
\end{tikzpicture}
\caption{\label{1D example}  Left panel: regression from data with asymmetric errors. 
Data are shown in blue; true model in black;  linear regression estimate in blue dash, and 
the new auto-tuned quantile estimate in red dash. Right: graph of $\log(1/\tau + 1/(1-\tau))$ in~\eqref{eq:QuantileEstimator}. 
This term, derived in Section~\ref{sec:model}, acts as a barrier pushing the quantile estimator into $(0,1)$. }
\end{figure}

Consider Figure \ref{1D example}, where linear observations have been 
contaminated with asymmetric errors, i.e. they are more likely to be positive than negative.
%\begin{wrapfigure}{r}{0.5\textwidth}
%\centering
%\end{wrapfigure}
The data generating mechanism is shown in black. The linear regression model for the data $\{y_i, a_i\}$   
 is equivalent to the least square problem,
\[\min_x \frac{1}{2}\|Ax - y\|^2\]
where $x$ contains slope and intercept. As expected, classic regression fails to capture the true mechanism because the errors are asymmetric, see 
blue dash in Figure~\ref{1D example}. 

In this case, we can actually recover the `true' line by assuming the errors arise from an asymmetric generalization of the Laplace distribution:
\begin{equation}
\label{eq:asymStat}
y_i = \langle a_i, x \rangle + \epsilon_i, \quad \epsilon_i \sim \exp(-\rho_\tau(\cdot)),
\end{equation}
where $\tau\in[0,1]$ controlling the relative slopes to the left and right of $0$, see Figure~\ref{fig:Qhub}(b).
Solving for a {\it joint} maximum likelihood estimator for $(x,\tau)$ , we
get the fit in red dash in Figure \ref{1D example}. 
The estimator is derived by considering the density corresponding to~\eqref{eq:asymStat}, and has the simple expression  
\begin{equation}
\label{eq:QuantileEstimator}
\min_{x, \tau \in [0,1]} q_{\tau} (Ax-b) + m\log\left(\frac{1}{\tau} + \frac{1}{1-\tau}\right).
\end{equation}
where the last term arises from considering the normalization constant required to make $\exp(-\rho_\tau(\cdot))$ a true statistical density, 
and is derived in Section~\ref{sec:model}, where the general approach is developed.  
In this simple example, the right solution is found by solving the single problem~\eqref{eq:QuantileEstimator} 
once, rather than considering multiple problem instances as we expect from classic approaches. 
The optimization problem itself is nonsmooth and nonconvex, but as we will see straightforward. 
The main thrust of the paper is to get a systematic way of formulating estimators such as~\eqref{eq:QuantileEstimator}, 
understand their properties, and consider different algorithms for solving these problems.

%\end{wrapfigure}

%Current approaches find $\tau$ by cross-validation or black-box optimization, where multiple 
%asymmetric regression problems with different $\tau$ must be solved, and a $\tau$ is then selected based on a validation criterion 
%(e.g. prediction accuracy on a held-out dataset). {\bf Our main contribution} is to fit both model and penalty parameters using an extended formulation. These extended problems are often {\it nonsmooth and nonconvex}, even in the case of the  quantile penalty in Figure~\ref{1D example}. {\bf Our second contribution} is to develop new optimization techniques for these problems.  In practice, the proposed approach is nearly as efficient as solving a single instance of the learning problem with fixed penalty parameters. 

\subsection{Contributions and Roadmap}
Building on the toolkit of PLQ penalties and corresponding densities developed in~\cite{JMLR:v14:aravkin13a}, we use the bridge between 
penalties and densities to develop an extended likelihood formulation over both shape parameters $\theta$ and regression variables $x$. 
The key idea is encoded in the {\it normalization constant}, that arises from the statistical interpretation and essentially acts 
as a balance against the data to ensure the final shape parameters still yield a statistically valid model. 
We consider properties of the extended likelihood formulations over $(x,\theta)$. In many cases, we show that 
standard techniques (including proximal alternating minimization and variable projection) can be applied. 
We also build on the interior point method of~\cite{JMLR:v14:aravkin13a} to develop a new extended interior point approach tailored to
these joint estimators.

The paper proceeds as follows. In Section~\ref{sec:QS} we review PLQ functions, along with their conjugate representations, and examples. 
Conjugate representations are essential for solving PLQ models using interior point methods.  
In Section~\ref{sec:model}, we focus on the statistical interpretation, and derive the normalization constant needed to create joint estimators such as~\eqref{eq:QuantileEstimator}. 
In Section~\ref{sec:algo}, we consider optimization methods that can be applied to the new class of estimators, and derive 
a new customized interior point method for the class. In Section~\ref{sec:synthetic} we provide synthetic verifications that show we
can recover shape parameters and regression variables in simple examples. Section~\ref{sec:real} contains some applications of these 
ideas to real datasets, particularly focusing on self-tuning penalties for robust PCA.

\section{Quadratic Support Functions and Their Conjugate Representations}
\label{sec:QS}

Piecewise linear-quadratic (PLQ) functions~\cite[Definition 10.20]{rockafellar2009variational} 
are convex functions whose domain can be represented as the union of finitely many polyhedral sets, relative to each of which 
the function can be written as a convex quadratic. These functions have a convenient representation using their convex conjugates.
\begin{definition}[Convex conjugate]
The {\it convex conjugate} of a function $f$ is given by 
 \[
 f^*(v) = \sup_{u} u\T v - f(u). 
 \]
\end{definition}

A PLQ function is defined as the conjugate to a quadratic over a polyhedral set~\cite[Example 11.18]{rockafellar2009variational}.
%\[
%p(x) = \sup_{Au \leq a} \left\{ \langle u, x\rangle  -\frac{1}{2}\langle u, Bu \rangle\right\}
%\]
Examples of common penalties used in statistical modeling, machine learning, and inverse problems can all be written using simple conjugates,
as shown in Table~\ref{tab:conj}.
Quadratic support (QS) functions~\cite{JMLR:v14:aravkin13a} generalize PLQ functions generalize this class by removing the polyhedral set restriction~\cite{JMLR:v14:aravkin13a}.

\begin{table}[h!]
\begin{tabular}{ccc}
Name & Conjugate Representation & Figure\\ \hline
Huber & 
\(h_\kappa(x) = \sup_{u\in[-\kappa, \kappa]}\left\{ ux - \frac{1}{2}u^2\right\}\) & 
Fig.~\ref{fig:Qhub}(a) \\ \hline
Quantile & 
 \(q_\tau(x) = \sup_{u \in [-\tau,(1-\tau)]} \left\{ux\right\}\) &
 Fig.~\ref{fig:Qhub}(b) \\ \hline
Quantile Huber &
\(h_{\tau, \kappa}(x) =  \sup_{u \in [-\kappa\tau,\kappa(1-\tau)]} \left\{ux - \frac{1}{2}u^2\right\}\) &
Fig.~\ref{fig:Qhub}(c) \\ \hline
Least squares & 
\(
\frac{1}{2}x^2 = \sup_{u} \left\{ux - \frac{1}{2}u^2\right\}
\)
& Fig.~\ref{fig:SDRex}(a)  \\ \hline
 Hinge & 
 \(h_\epsilon(x) = \sup_{u \in [-\tau,(1-\tau)]} \left\{u(x-\epsilon)\right\}\) &
 Fig.~\ref{fig:SDRex}(b) \\ \hline
Vapnik & 
\(
\rho_\epsilon(x) = \sup_{u \in [0,1]^2}
\left\{
\left\langle 
\begin{bmatrix}1  \\ 
-1 \end{bmatrix}x 
-
\begin{bmatrix}\epsilon \\ 
\epsilon \end{bmatrix} , 
u \right\rangle
\right\}
\) &
Fig.~\ref{fig:SDRex}(c) \\ \hline
Smooth insensitive loss  &
\(
\rho^h_\epsilon(x) = \sup_{u \in [0,1]^2}
\left\{
\left\langle 
\begin{bmatrix}1  \\ 
-1 \end{bmatrix}x 
-
\begin{bmatrix}\epsilon \\ 
\epsilon \end{bmatrix} , 
u \right\rangle - \frac{1}{2}u^Tu
\right\} 
\)  &
Fig.~\ref{fig:SDRex}(d) \\ \hline
Elastic net & 
\(
\rho(x) = 
\sup_{u \in [0,1] \times \mathbb{R}}
\left\{
\left\langle 
\begin{bmatrix}1  \\ 
1 \end{bmatrix}x , u \right\rangle
 - \frac{1}{2}u^T \begin{bmatrix} 0 & 0 \\ 0 & 1\end{bmatrix}u\right\}
\) &
Fig.~\ref{fig:SDRex}(e) \\\hline
%Hybrid &
%\(
%h_{\epsilon}(x) = \sup_{u \in \left[-\epsilon^{-1}, \epsilon^{-1} \right]} \left\{xu - \left(1-\sqrt{1-(u\epsilon)^2}\right)\right\}
%\) &
%Fig.~\ref{fig:SDRex}(b) \\ \hline
%Logistic & 
%\(
%h_a(x) = \sup_{u \in [0,a]} \left\{xu - \frac{u}{a}\log\left(\frac{u}{a}\right) -
%\left(1-\frac{u}{a}\right)\log\left(1-\frac{u}{a}\right)\right\}
%%\end{equation}
%\) &
%Fig.~\ref{fig:SDRex}(c)
\end{tabular}
\caption{\label{tab:conj}Common PLQ Functions and their Conjugate Representations}
\end{table}

%Note that while Definition~\ref{def:SDR} does not require $C$ 
%to be sparse, this property holds for all known examples; this feature will be important for computation in later sections. 
\begin{figure}[h!]
\center
\begin{tabular}{ccccc}\\ 
%%%%%%%%%%%%%%%%%%%%%%%%%%%%%%%%%%%%%%%%%%%%%%%%
% L2 graph %%%%%%%%%%%%%%%%%%%%%%%%%%%%%%%%%%%%%%%%%%
%%%%%%%%%%%%%%%%%%%%%%%%%%%%%%%%%%%%%%%%%%%%%%%%
\begin{tikzpicture}
  \begin{axis}[
    thick,
    height=2cm,
    xmin=-2,xmax=2,ymin=0,ymax=1,
    no markers,
    samples=50,
    axis lines*=left, 
    axis lines*=middle, 
    scale only axis,
    xtick={-1,1},
    xticklabels={},
    ytick={0},
    ] 
%\addplot[red,domain=-2:-1,densely dashed]{-x-.5};
\addplot[blue, domain=-2:+2]{.5*x^2};
%\addplot[red,domain=+1:+2,densely dashed]{x-.5};
%\addplot[blue,mark=*,only marks] coordinates {(-1,.5) (1,.5)};
  \end{axis}
  \end{tikzpicture}
%&\begin{tikzpicture}
%  \begin{axis}[
%    thick,
%    height=2cm,
%    xmin=-2,xmax=2,ymin=0,ymax=1,
%    no markers,
%    samples=100,
%    axis lines*=left, 
%    axis lines*=middle, 
%    scale only axis,
%    xtick={-1,1},
%    xticklabels={},
%    ytick={0},
%    ] 
%\addplot[blue-violet, domain=-2:+2]{sqrt(1+(x/1)^2) - 1};
%  \end{axis}
%\end{tikzpicture}
%&
%\begin{tikzpicture}
%  \begin{axis}[
%    thick,
%    height=2cm,
%    xmin=-2,xmax=2,ymin=0,ymax=1,
%    no markers,
%    samples=100,
%    axis lines*=left, 
%    axis lines*=middle, 
%    scale only axis,
%    xtick={-1,1},
%    xticklabels={},
%    ytick={0},
%    ] 
%\addplot[brightturquoise, domain=-2:+2]{ln(1+exp(x-1)};
%  \end{axis}
%\end{tikzpicture}
&\begin{tikzpicture}
  \begin{axis}[
    thick,
    height=2cm,
    xmin=-2,xmax=2,ymin=0,ymax=1,
    no markers,
    samples=50,
    axis lines*=left, 
    axis lines*=middle, 
    scale only axis,
    xtick={-1,1},
   xticklabels={},
    ytick={0},
    ] 
\addplot[red,domain=-2:0.5,densely dashed]{0*x};
\addplot[red,domain=0.5:+2,densely dashed]{x-.5};
%\addplot[blue,mark=*,only marks] coordinates {(-1,.5) (1,.5)};
  \end{axis}
\end{tikzpicture}
% \begin{tikzpicture}
%  \begin{axis}[
%    thick,
%    height=2cm,
%    xmin=-2,xmax=2,ymin=0,ymax=1,
%    no markers,
%    samples=100,
%    axis lines*=left, 
%    axis lines*=middle, 
%    scale only axis,
%    xtick={-.24,.56},
%    xticklabels={},
%    ytick={0},
%    ] 
%\addplot[red,domain=-2:-2*0.3*0.4,densely dashed]{0.3*abs(x) - 0.4*0.3^2};
%\addplot[blue,domain=-2*0.3*0.4:2*(1-0.3)*0.4]{0.25*x^2/0.4};
%\addplot[red,domain=2*(1-0.3)*0.4:2,densely dashed]{(1-0.3)*abs(x) - 0.4*(1-0.3)^2};
%\addplot[blue,mark=*,only marks] coordinates {(-.24,0.0550) (0.56,0.20)};
%  \end{axis}
%\end{tikzpicture} \\ 
& \begin{tikzpicture}
  \begin{axis}[
    thick,
    height=2cm,
    xmin=-2,xmax=2,ymin=0,ymax=1,
    no markers,
    samples=50,
    axis lines*=left, 
    axis lines*=middle, 
    scale only axis,
    xtick={-0.5,0.5},
    xticklabels={},
    ytick={0},
    ] 
    \addplot[red,domain=-2:-0.5,densely dashed] {-x-0.5};
    \addplot[domain=-0.5:+0.5] {0};
    \addplot[red,domain=+0.5:+2,densely dashed] {x-0.5};
    \addplot[blue,mark=*,only marks] coordinates {(-0.5,0) (0.5,0)};
  \end{axis}
\end{tikzpicture}
& \begin{tikzpicture}
  \begin{axis}[
    thick,
    height=2cm,
    xmin=-2,xmax=2,ymin=0,ymax=1,
    no markers,
    samples=50,
    axis lines*=left, 
    axis lines*=middle, 
    scale only axis,
    xtick={-1,1, -.5, .5},
    xticklabels={},
    ytick={0},
    ] 
\addplot[domain=-0.25:+0.25] {0};
\addplot[red,domain=-2:-1,densely dashed]{-x-.5-.5*.25};
\addplot[blue, domain=-1:-.25]{.5*x^2-.5*.25};
\addplot[blue, domain=.25:1]{.5*x^2-.5*.25};
\addplot[red,domain=+1:+2,densely dashed]{x-.5-.5*.25};
\addplot[blue,mark=*,only marks] coordinates {(-1,.5-.5*.25) (1,.5-.5*.25)(-.45, 0) (.45, 0)};
  \end{axis}
\end{tikzpicture}
& \begin{tikzpicture}
  \begin{axis}[
    thick,
    height=2cm,
    xmin=-2,xmax=2,ymin=0,ymax=1,
    no markers,
    samples=100,
    axis lines*=left, 
    axis lines*=middle, 
    scale only axis,
    xtick={-1,1},
    xticklabels={},
    ytick={0},
    ] 
\addplot[amethyst, domain=-2:+2]{.5*x^2 + 0.5*abs(x)};
  \end{axis}
\end{tikzpicture}
\\ 
(a) quadratic
%& (b)  hybrid loss, $\epsilon = 1$
%&(c) logistic loss, $a=2$
&(b)  hinge, $\epsilon = 0.5$
%&(c) quantile Huber\\ 
&(c) Vapnik, $\epsilon = 0.5$ 
&(d) Huber insensitive loss 
& (e) elastic net ($\alpha = 0.5)$
\end{tabular}
\caption{\label{fig:SDRex} Eight QS penalties frequently used in machine learning. Huber and quantile losses are shown in Fig.~\ref{fig:Qhub}.}
\end{figure}

PLQ functions are closed under sums and affine compositions~\cite{JMLR:v14:aravkin13a}, and we can explicitly define 
PLQ functions using their conjugate representation.  
%the conjugate of $\frac{1}{2}u^TMu + \delta_{U}(u)$, evaluated at point $Br-\bar b$~\cite{JMLR:v14:aravkin13a}:

\begin{definition}[PLQ Functions and penalties.]
A piecewise linear-quadratic (PLQ) function is given by 
\begin{equation}
\label{eq:QSdual}
%&\rho(r;B,\bar{b},C,\bar{c},M) \\
\hspace{-.1in}
\rho(r) = \sup_{u} \left\{u\T(Br-\bar{b}) - \frac{1}{2} u\T Mu : C\T u\le \bar{c}\right\},
\end{equation}
where $M\succeq0$, $B$ is an injection, and $U:=\{u : C\T u\le \bar{c}\}$ is a polyhedral. 

When $0 \in U$, we may call $\rho$ a {\it penalty} to emphasize that it is non-negative.  
\end{definition}

The notion of {\it coercivity} is essential to the statistical interpretation of PLQs in the next section. 

\begin{definition}[Coercivity]
A function $f(x)$ is coercive if 
\[
\lim\inf_{|x|\rightarrow \infty} f(x) = \infty.
\] 
\end{definition}
Geometric conditions for coercivity of PLQs are given in~\cite[Theorem 10]{JMLR:v14:aravkin13a}. In the next section, we use PLQs and their properties 
to build a bridge to densities, which enable the joint likelihood approach.

%======================================================================
\section{Statistical Model and Properties of Joint Objective}
\label{sec:model}
To formulate a learning problem over spaces of penalties, we 
first review the relationship between penalties and corresponding residual distributions. 
%{error distributions. - should we call it residual distribution?}
We then use this relationship to develop a {joint} formulation for {model and} shape parameter inference, 
and characterize properties of the resulting objective function. 

%For maximum clarity to illustrate the idea, we consider linear regression through section 2, 3 and 4.
%However, our work is definitely not only restricted to linear regression problems.
%In section 5 and 6 we study the robust PCA and auto-encoder as concrete extensions of our model.
%\subsection{Statistical view}
%Recall the quantile penalty in Figure~\ref{fig:Qhub}. 
%If we choose $\tau$ to be close to 1, we penalize the negative errors a lot more than the positive.
%This is equivalent to {assuming} that the errors $\epsilon_i$ {have a positive bias.}
%is biased towards positive.
%errors.  
%It is equivalent to the assumption which errors come from a distribution that is more likely positive than negative.
%\vskip 8pt

The relationship between penalties and associated densities can be made precise. Every coercive PLQ penalty is integrable. 
Given a coercive penalty $\rho(r; \theta)$ whose shape is parametrized by $\theta$, we define an associated density
\begin{equation}
\label{eq:penalty-density}
\begin{aligned}
p(r;\theta) &= \frac{1}{n_c(\theta)}\exp[-\rho(r;\theta)],  \quad \text{where} \quad n_c(\theta) &= \int_\mathbb{R} \exp[-\rho(r;\theta)]\,dr.
\end{aligned}
\end{equation}
The term $n_c(\theta)$ is a normalization constant that ensures that $\rho(r;\theta)$ in~\eqref{eq:penalty-density} is a true statistical density, 
i.e. it integrates to $1$. 
These observations yield a simple recipe for extending a negative log likelihood in $x$ (based on optimizing a PLQ penalty) to 
inform shape parameters $\theta$. Specifically, given an optimization problem of the form 
\[
\min_x \sum_{i=1}^m \rho(y_i - \ip{a_i,x};\theta) 
\]
we consider the extended problem 
\begin{equation}
\label{eq:obj}
\min_{x,\theta\in\mathcal{D}} \sum_{i=1}^m \rho(y_i - \ip{a_i,x};\theta) + m\log[n_c(\theta)],
\end{equation}
where $\theta$ may be restricted to a domain $\mathcal{D}$. 
For example, in the quantile regression problem in Section~\ref{ex:quantileReg}, we have  $\mathcal{D} = [0,1]$ and 
\[
\rho_\tau(x) = \begin{cases}
(1-\tau)x & x \geq 0 \\
-\tau x & x \leq 0
\end{cases}, 
\quad p(r; \tau) = \begin{cases}
\exp(-(1-\tau)x) & x \geq 0 \\
\exp(-\tau x) & x \leq 0
\end{cases}, 
\quad n_c(\tau) = \frac{1}{1-\tau} + \frac{1}{\tau}.
\]
The $\log(n_c)$ term acts as a barrier (see Figure~\ref{1D example}), pushing $\tau$ away from the boundary points $0$ and $1$ into the interior $(0,1)$, 
and  favoring $\tau = 0.5$. %violating a key assumption often required by optimization algorithms. %The joint objective is $(x,\theta)$ is nonsmooth and nonconvex. 

Interestingly, the log of the normalization constant for the quantile regression problem is smooth and strongly convex, but its gradient is not Lipschitz continuous.
In the remainder of this section, we characterize theoretical properties of the general objective~\eqref{eq:obj}. 

\begin{assumption}
\label{asp:smoothness}
To ensure the validity of the statistical viewpoint, we require $\rho$ to satisfy:
\begin{enumerate}
\item $\rho(r;\theta)\ge0$ is a PLQ {\it penalty} for every $\theta\in\cD$, so we have  \textbf{non-negativity}.
\item $\rho$ is {\it coercive} for any $\theta\in\mathcal{D}$, so we have \textbf{integrability}.
\item For any $\theta_0\in\mathcal{D}$, $\rho(r;\theta)$ is $C^2$ around $\theta_0$ for almost every $r\in\mathbb{R}$ (\textbf{smoothness in $\theta$})
\end{enumerate}
\end{assumption}
Under these assumptions, we can obtain formulas for the first and second derivatives of $n_c(\theta)$. 
\begin{theo}[smoothness of $n_c(\theta)$]
\label{th:smoothness}
For $n_c(\theta)$in~\eqref{eq:penalty-density}, suppose Assumption \ref{asp:smoothness} holds 
and for $\theta_0\in\mathcal{D}$, there exist functions $g_k(r)$, $k=1,2$, such that,
\begin{enumerate}
\item for any unit vector $v$, $|\ip{\nabla_\theta\exp[-\rho(r;\theta)],v}|\le g_1(r)$ for any $\theta$ around $\theta_0$,
\item for any unit vector $v$, $\left|\ip{\nabla_\theta^2\exp[-\rho(r;\theta)]v,v}\right|\le g_2(r)$ for any $\theta$ around $\theta_0$,
\item $\int_\mathbb{R} g_k(r)\,dr < \infty$, $k=1,2$.
\end{enumerate}
then $n_c(\theta)$ is $C^2$ continuous around $\theta_0$ and,
\begin{equation}
\label{eq:nc_form}
\begin{aligned}
\nabla n_c(\theta_0) &= \int_\mathbb{R} \nabla_\theta\exp[-\rho(r;\theta_0)]\,dr, \\ 
\nabla^2 n_c(\theta_0) &= \int_\mathbb{R}\nabla_\theta^2\exp[-\rho(r;\theta_0)]\,dr.
\end{aligned}
\end{equation}
\end{theo}
The proof is elementary, and is included below for completeness. 
\begin{proof}
From Assumption \ref{asp:smoothness}, we know that for any $\theta_0\in\mathcal{D}$, $\nabla_\theta\exp[\rho(r;\theta_0)]$ and $\nabla_\theta^2\exp[\rho(r;\theta_0)]$ exist for almost every $r\in\mathbb{R}$. For any $h$ such that $\|h\|$ is small enough to make $\theta_0 + h$ stay in the neighborhood of $\theta_0$. By applying mean value theorem, we have,
\begin{align*}
n_c(\theta_0+h) - n_c(\theta_0) & = \int_\mathbb{R} \exp[-\rho(r;\theta_0 + h)] - \exp[-\rho(r;\theta_0)]\,dr
= \int_\mathbb{R} \ip{\nabla_\theta\exp[-\rho(r;\bar{\theta})],h}\,dr  \\
 \Rightarrow &\frac{n_c(\theta_0+h) - n_c(\theta_0)}{\|h\|}
= \int_\mathbb{R} \ip{\nabla_\theta\exp[-\rho(r;\bar{\theta})],\frac{h}{\|h\|}}\,dr
\end{align*}
where $\bar{\theta}$ lie in segment with the end points $\theta_0$ and $\theta_0 + h$. The first and third assumptions allow us to apply the dominant convergence theorem and get,
\begin{align*}
&\lim_{h\rightarrow0}\frac{n_c(\theta_0+h) - n_c(\theta_0)}{\|h\|} = \lim_{h\rightarrow0}\int_\mathbb{R} \ip{\nabla_\theta\exp[-\rho(r;\bar{\theta})],\frac{h}{\|h\|}}\,dr
= \int_\mathbb{R} \lim_{h\rightarrow0}\ip{\nabla_\theta\exp[-\rho(r;\bar{\theta})],\frac{h}{\|h\|}}\,dr\\
&= \int_\mathbb{R} \ip{\nabla_\theta\exp[-\rho(r;\theta_0)],v}\,dr = \ip{\int_\mathbb{R}\nabla_\theta\exp[-\rho(r;\theta_0)]\,dr,v}
\end{align*}
where we set $h = \alpha v$ and let $\alpha \rightarrow 0^+$ and keep $v$ fix as an unit vector. From the definition of the gradient we know that,
\[\nabla n_c(\theta_0) = \int_\mathbb{R}\nabla_\theta\exp[-\rho(r;\theta_0)]\,dr.\]
Following the same steps we could also show $\nabla^2c(\theta_0)$ exists and satisfies,
\[\nabla^2 n_c(\theta_0) = \int_\mathbb{R}\nabla_\theta^2\exp[-\rho(r;\theta_0)]\,dr.\]
\end{proof}

The derivative formulas~\eqref{eq:nc_form} are used for first- and second-order methods to infer $x$ and $\theta$. The parametrization conditions in $\theta$ are satisfied by all piecewise linear-quadratic (PLQ) penalties.  The theorem can also be applied to densities that are not log-concave. For {instance}, the Student's $t$ density  
%\footnote{
%\[
%p(r, \nu) = \frac{\Gamma(\nu + 1)/2}{\sqrt{\pi \nu}\Gamma(\nu/2)}\left(1 + \frac{r^2}{\nu}\right)^{-(\nu+1)/2}.
%\]}
and associated penalty satisfy all assumptions of Theorem~\ref{th:smoothness}. % for $\nu > 1$. 
\begin{figure}[h]
\centering
\includegraphics[width=8cm]{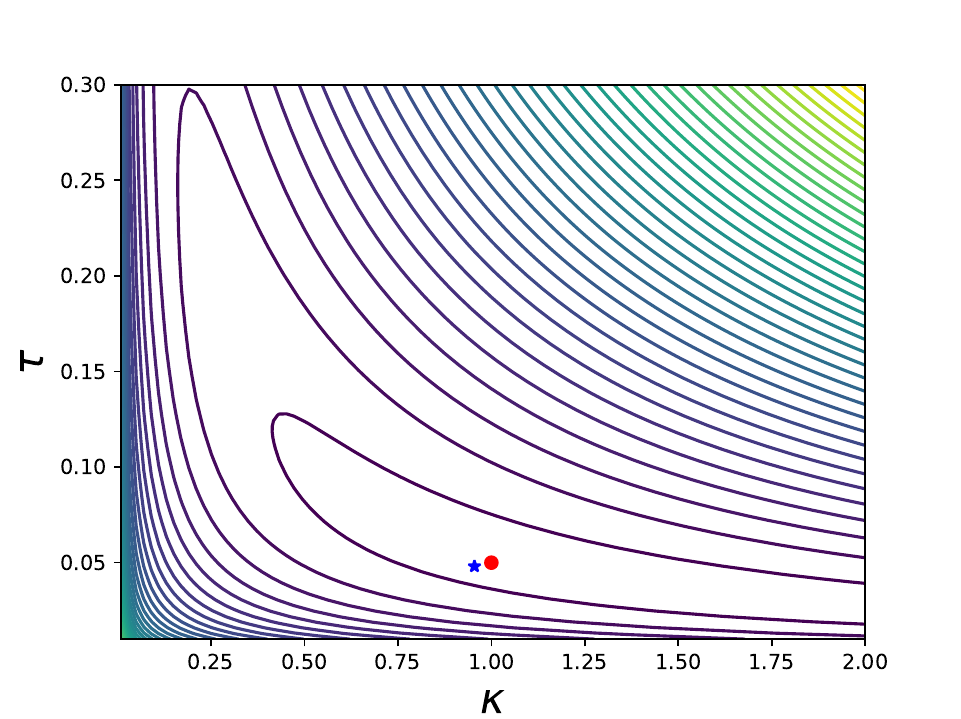}
\caption{\small\label{fig:levelset} Level sets for value function $v(\theta)$~\eqref{eq:value} for the quantile Huber model. 
The blue star is the maximum likelihood estimator, while the red dot represents the true parameters in the simulation.}
\end{figure}

In the quantile case~\eqref{eq:QuantileEstimator}, the term $\log[n_c(\theta)]$ is convex. We characterize sufficient conditions 
for convexity of $\log[n_c(\theta)]$ for a general class of penalties $\rho$. The results can be derived using \cite[Chapter 3.5]{boyd2004convex}.  
\begin{theo}[convexity of $\log\lbrack n_c(\theta)\rbrack$]
\label{th:convexity}
Let $n_c(\theta)$ be defined as in Theorem \ref{th:smoothness}, and suppose Assumption \ref{asp:smoothness} holds. We have the following results:
\begin{enumerate}
\item If $\rho(r;\theta)$ is jointly convex in $r$ and $\theta$, then $\log[n_c(\theta)]$ is a concave function of $\theta$.
\item If $\rho(r;\theta)$ is concave with respect to $\theta$ for every $r$, then $\log[n_c(\theta)]$ is a convex function.
\end{enumerate}
\end{theo}
%\begin{proof} please check the Appendix A for the proof. \end{proof}

Theorems~\ref{th:smoothness} and~\ref{th:convexity} tell an interesting story. The log-normalization constant 
$\log[n_c(\theta)]$ is nearly always smooth; even when the loss $\rho$ is nonsmooth in $x$. 
The inference problem~\eqref{eq:obj} is {\it never guaranteed to be jointly convex} in $(x,\theta)$:  
if $\rho(x;\theta)$ is jointly convex, then $\log[n_c(\theta)]$ will be {\it concave}. This is intuitive, as 
we are attempting to learn both the model and error structure at the same time. Objective~\eqref{eq:obj} is generally   nonsmooth and nonconvex. In the next section, we show how to optimize it using 
first and second order methods.

\subsection{Level sets of the objective function and maximum likelihood estimate}
\label{sec:QHexample}
Although \eqref{eq:obj} is non-convex, in {many} cases we can still find the global minimum in $\theta$. To illustrate, we generate the samples $\epsilon_i$ from distribution defined by quantile Huber function with $\kappa = 1$ and $\tau = 0.05$, select a set of weights $x$, generate data $y_i = \ip{a_i, x} + \epsilon_i$, 
and plot the so called {\it value function} for \eqref{eq:obj}:
\begin{equation}
\label{eq:value}
v(\theta) = \min_{x} \sum_{i=1}^m \rho(y_i - \ip{a_i,x};\theta) + m\log[n_c(\theta)].
\end{equation}
Evaluating $v(\theta)$ requires solving a smooth convex problem in $x$, and thus 
problem~\eqref{eq:obj} reduces to minimizing~\eqref{eq:value} in $\theta$.  
Results for the simple 2D case are shown in Figure \ref{fig:levelset}. $v(\theta)$ is non-convex (note the level sets), but there is a unique minimum that is close to the true parameters, and was found in every run by a local search. 
In the next section, we design methods that are far more efficient than optimizing $v(\theta)$.

%{\bf Remark}
%\begin{itemize}
%\item \eqref{eq:obj} could be non-convex and non-smooth depend on the $\rho$ you choose.
%\item There is no bi-convexity either, namely when we fix $x$ or $\theta$ \eqref{eq:obj} may not be convex or concave.
%\item There are some special case that $\eqref{eq:obj}$ is bi-convex, e.g., quantile penalty.
%\item In general, we need algorithms that could deal with non-convex non-smooth functions.
%\end{itemize}

%======================================================================
\section{Optimization Methods for Joint PLQ and Shape Parameter Estimation.}
\label{sec:algo}

In this section, we discuss algorithms for~\eqref{eq:obj}, and develop a new interior point method building on the PLQ 
optimization strategy of~\cite{JMLR:v14:aravkin13a}
When $\rho$ is smooth in $x$ and $\theta$, we show how to apply  
Proximal Alternating Linearized Minimization (PALM)~\cite{bolte2014proximal} and Proximal Alternating Minimization (PAM)~\cite{attouch2010proximal}.
We also discuss variable projection (VP) type algorithms~\cite{aravkin2012estimating,aravkin2017efficient}.
These development is straightforward, but the log normalization constant $\log[n_c(\theta)]$
must be treated carefully, as its gradient does not have a global Lipschitz constant.

Requiring smoothness in $\rho$ is restrictive, and the quantile regression example~\eqref{eq:quantile} has a nonsmooth $\rho$.
The quantile penalty is not smooth, but it is PLQ. We  use conjugate representations of PLQs in Section~\ref{sec:QS} 
to extend interior point methods developed in~\cite{JMLR:v14:aravkin13a} to problem~\eqref{eq:obj}
in Section~\ref{sec:IP}. 

\subsection{PALM, PAM, and Variable Projection}
\label{sec:PALM}

Several algorithms from existing literature can be brought to bear on problem~\eqref{eq:obj} when it is of the form  
\begin{equation}
\label{eq:PalmForm}
\min_{x,\theta} H(x,\theta) + r_1(x) + r_2(\theta),
\end{equation}
under some mild assumptions on $H$. Here we briefly review PALM~\cite{bolte2014proximal}, and PAM~\cite{attouch2010proximal} algorithms, 
as well as similar algorithms that partially minimize the objective in $\theta$, known as {\it partial minimization} or {\it variable projection}~\cite{aravkin2012estimating,aravkin2017efficient}.

The PALM and PAM algorithms \cite{bolte2014proximal} can be used to minimize~\eqref{eq:PalmForm} when 
$H$ is $C^1$, with globally Lipschitz partial gradients, while the functions $r_1$ and $r_2$ are proper lower semicontinuous; 
in particular they are not required to be convex, finite-valued, or smooth. 
The term $r_1(x)$ is useful if we need simple regularization and constraints on $x$, such as sparsity or non-negativity. 
Even though 
$\log[n_c(\theta)]$ is smooth (see Theorem~\ref{th:smoothness}), it must be relegated to $r_2(\theta)$, 
since otherwise it violates the Lipschitz assumptions on the partial gradients of $H$. Therefore, to apply PALM or PAM to~\eqref{eq:obj}, 
we take 
\begin{equation}
\label{eq:palm_detail}
\begin{aligned}
H(x,\theta) &= \sum_{i=1}^m \rho(y_i - \ip{a_i,x};\theta), \qquad r_2(\theta) = \delta_\mathcal{D}(\theta) + m\log[n_c(\theta)].
\end{aligned}
\end{equation}
Here $\delta_{\mathcal{D}}$ is the indicator function for the set $\mathcal{D}$, 
\[
\delta_{\mathcal{D}} (\theta) = \begin{cases} 0 & 
\mbox{if } \theta \in \mathcal{D} \\ \infty & \mbox{if } \theta\not\in    \mathcal{D} \end{cases}.
\]
The PALM algorithm is detailed in Algorithm~\ref{alg:PALM}.
The steps $c_k$ and $d_k$ are obtained from Lipschitz constants of the (partial) gradients of $H$.
The PAM algorithm is given in Algorithm~\ref{alg:PAM}, with the same step sizes to facilitate an easier comparison. 

PAM essentially requires the ability to partially minimize a quadratically regularized problem $H$ with respect to 
both $x$ and $\theta$. In many cases, one of these variables may be easier to solve for than the other; 
in Example~\ref{sec:QHexample}, $\theta$ has dimension $2$. Variable projection algorithms~\cite{aravkin2012estimating,aravkin2017efficient} 
exploit the ability to fully minimize in one variable while applying an iterative approach in the other. 
Algorithm~\ref{alg:VP} shows an example where we have chosen to optimize out $\theta$ for every iteration in $x.$

%PALM algorithm requires $H$ to be smooth and the partial gradient of $H$ to be Lipschitz continuous.
%But $\log[n_c(\theta)]$ usually do not have a quadratic upper bound for $\theta$ which make PALM not feasible.
\begin{algorithm}
\caption{PALM for~\eqref{eq:palm_detail}}\label{alg:PALM}
\begin{algorithmic}[1]
\Require{$A$, $y$}
\State \textbf{Initialize}: $x^0$, $\theta^0$ %, $k\leftarrow 0$
\While{not converge}
%\Let{$c_k$}{$\gamma_1L_1(\theta^k)$}
\Let{$x^{k+1}$}{$\prox_{\frac{1}{c_k}r_1}\left(x^k - \frac{1}{c_k}\nabla_x H(x^k,\theta^k)\right)$}
%\Let{$d_k$}{$\gamma_2L_2(x^{k+1})$}
\Let{$\theta^{k+1}$}{$\prox_{\frac{1}{d_k}r_2}\left(\theta^k - \frac{1}{d_k}\nabla_\theta H(x^{k+1},\theta^k)\right)$}
%\Let{$k$}{$k+1$}
\EndWhile
\Return{$x^k$ and $\theta^k$}
\end{algorithmic}
\end{algorithm}

\begin{algorithm}
\caption{PAM for~\eqref{eq:palm_detail}}\label{alg:PAM}
\begin{algorithmic}[1]
\Require{$A$, $y$}
\State \textbf{Initialize}: $x^0$, $\theta^0$ %, $k\leftarrow 0$
\While{not converge}
%\Let{$c_k$}{$\gamma_1L_1(\theta^k)$}
\Let{$x^{k+1}$}{$\min_x H(x,\theta^k) + r_1(x) + \frac{1}{c_k}\|x - x^k\|^2$}
%\Let{$d_k$}{$\gamma_2L_2(x^{k+1})$}
\Let{$\theta^{k+1}$}{$ \min_\theta H(x^{k+1},\theta) + r_2(\theta) + \frac{1}{d_k}\|\theta - \theta^k\|^2$}
%\Let{$k$}{$k+1$}
\EndWhile
\Return{$x^k$ and $\theta^k$}
\end{algorithmic}
\end{algorithm}

\begin{algorithm}
\caption{VP for~\eqref{eq:palm_detail}, partially minimizing w.r.t. $\theta$}\label{alg:VP}
\begin{algorithmic}[1]
\Require{$A$, $y$}
\State \textbf{Initialize}: $x^0$, $\theta^0$ %, $k\leftarrow 0$
\While{not converge}
%\Let{$c_k$}{$\gamma_1L_1(\theta^k)$}
\Let{$x^{k+1}$}{$\prox_{\frac{1}{c_k}r_1}\left(x^k - \frac{1}{c_k}\nabla_x H(x^k,\theta^k)\right)$}
%\Let{$d_k$}{$\gamma_2L_2(x^{k+1})$}
\Let{$\theta^{k+1}$}{$\min_\theta H(x^{k+1},\theta) + r_2(\theta)$}
%\Let{$k$}{$k+1$}
\EndWhile
\Return{$x^k$ and $\theta^k$}
\end{algorithmic}
\end{algorithm}

{\bf Detail:} The prox operator of $\log[n_c(\theta)]$ is not available in closed form for any examples of interest. 
Instead, it can be efficiently computed using the results of Theorem~\ref{th:smoothness}: 
\begin{equation}
\label{eq:prox_compute}
\prox_{\frac{1}{d_k}r_2}(\phi) = \arg\min_{\theta\in \mathcal{D}} \frac{1}{2d_k}\|\theta - \phi\|^2 + \log[n_c(\theta)]. 
\end{equation}
In all examples of interest, $\theta$ is low dimensional; and we compute~\eqref{eq:prox_compute} using Newton's method
or an interior point method. 
This requires $\nabla \log[n_c(\theta)]$ and $\nabla^2 \log[n_c(\theta)]$, which are calculated  
numerically using formulas~\eqref{eq:nc_form}. 
%The cost of this step is negligible compared to 
%the evaluation $\nabla_x H(x,\theta)$. 

The PALM algorithm works well for large-scale shape inference problems with smooth $\rho$. 
We use it for the self-tuning RPCA experiments in Section~\ref{sec:real}. %\red{Do we want to call this large-scale?}

\subsection{Interior Point method for nonsmoothly coupled nonconvex joint QS inference}
\label{sec:IP}
%The restriction that $\rho$ must be smooth in $(x,\theta)$ is restrictive.
In this section, we use conjugate representations of PLQ penalties to develop an interior point method for the extended inference problem~\eqref{eq:obj}:
\[
\min_{x,\theta \in \mathcal{D}} \rho(x;\theta) + \log[n_c(\theta)].
\]
%adding shape parameter tuning. 
The approach converges {superlinearly} in practice, but each iteration requires solving a linear system. 
We solve these systems directly, and scaling issues are explored in Table~\ref{tb:runtime}. 
In practice, large-scale linear systems are solved iteratively, often using pre-conditioners~\cite{orban2017iterative};
we leave these developments for future work.

%The approach is limited to moderate problem dimensions\footnote{If  
% $A$ has dimensions $m$ and $n$, interior point methods require $O(n(m^2+n^2))$ arithmetic operations, 
% where $n$ is the smaller dimension. This limits practical applications for large-scale problems; to go beyond $2000 \times 2000$ with modest compute, some sort of special structure or technique (sparsity, preconditioning) is typically needed.}, but converges at a superlinear rate, and solves problems with nonsmooth coupling in $(x,\theta)$.

To optimize  PLQ penalties parametrized by $\theta$, we allow $\bar b$ and $\bar c$ to be affine 
functions of $\theta$,  and assume $\mathcal{D}$ is also polyhedral: 
\[
\bar{b} = G\T\theta + b,\quad\bar{c} = H\T\theta + c,
\quad
\mathcal{D} = \{\theta: S\T\theta\le s\}.
\]
We then solve a {saddle point} for primal variables $x$, conjugate 
variables $u$, and shape parameters $\theta$:
\begin{equation}
\label{eq:QSprimaldual}
\begin{aligned}
\min_{x,S\T\theta\le s} &\sup_{u} \left\{ u\T[B(Ax-y)-G\T\theta - b]-\frac{1}{2}u\T M u:  C\T u \le H\T\theta + c \right\} + m\log[n_c(\theta)]
\end{aligned}
\end{equation}
For example, the self-tuning quantile penalty~\eqref{eq:QuantileEstimator} is written as  
\[
\begin{aligned}
\min_{x, \tau \in \mathcal{C}_1} &
\quad \sup_{(u,\tau) \in \mathcal{C}_2}
\lt\{u\T(Ax-b) + m\log\left(\frac{1}{\tau} + \frac{1}{1-\tau}\right)\rt\}.\\
\mathcal{C}_1 &:= \left\{ \tau: \begin{bmatrix} 1 \\ -1 \end{bmatrix}\tau \leq \begin{bmatrix}1 \\ 0 \end{bmatrix}\right\}, \quad  \mathcal{C}_2  = \left\{(u, \tau): \begin{bmatrix}1 \\ -1 \end{bmatrix} u \leq -\begin{bmatrix}1 \\ 1\end{bmatrix} 
\tau +\begin{bmatrix} 
1 \\ 0\end{bmatrix}\right\}.
\end{aligned}
\]
\begin{algorithm}[h!]
\caption{Interior point method for QS with $\theta$ estimation}\label{alg:IPsolve}
\begin{algorithmic}[1]
\Require{$A$, $y$, $B$, $b$, $C$, $c$, $G$, $H$, $S$, $s$}
\State \textbf{Initialize}: $z^0$, $k\leftarrow0$, $\mu = 1$
\While{not converged}
\Let{$p$}{$\nabla F_\mu(z^k)^{-1}F_\mu(z^k)$}
\Let{$\alpha$}{$\text{LineSearch}(z^k,p)$, using $\|F_\mu(\cdot)\|$}
\Let{$z^{k+1}$}{$z^k - \alpha p$}
%\Let{$k$}{$k + 1$}
\Let{$\mu$}{$0.1\cdot$ (Average complementarity conditions)}
\EndWhile
\Return{$z^{k+1}$}
\end{algorithmic}
\end{algorithm} Interior point (IP) methods apply damped Newton to a relaxation of the optimality conditions {of}~\eqref{eq:QSprimaldual}, see~\cite{KMNY91,NN94,Wright:1997}. % \red{sasha to include references}.
Let $F(z)$ denote of optimality conditions for problem~\eqref{eq:QSprimaldual}, with $z=(x,u,\theta, \lambda)$, where 
$\lambda$ are dual variables for inequality constraints $C\T u \le H\T\theta + c$ and $S\T\theta\le s$, 
and let $F_\mu(z)$ denote the relaxed system obtained 
by using a logarithmic barrier with parameter $\mu$. The IP method is summarized in Algorithm \ref{alg:IPsolve}.
%with  details in the supplementary materials. 
% The barrier parameter $\mu$ is taken to $0$ 
%as a specific fraction of the average complementarity conditions in $F(z)$.

We introduce a log-barrier for the conjugate variables $u$ and shape parameters $\theta$ as follows. 
\begin{align*}
&\delta_{\mathcal{D}}(\theta) \approx -\mu\one\T\log(s-S\T\theta),\\
&\delta_{\{(u,\theta)\mid C\T u \le H\T\theta + c\}}(u,\theta) \approx -\mu\one\T\log(c + H\T\theta-C\T u).
\end{align*}
As  $\mu \downarrow 0$, the barriers approach true indicator functions for the $\mathcal{D}$ and $U$. 
The parameter $\mu$ is decreased to a specified tolerance as the optimization proceeds. 
For fixed $\mu$, the approximate subproblem with fixed $\mu$ can itself be written as a saddle point system: 
\begin{equation}
%\tag{S.1}
\begin{aligned}
\label{eq:QSprimaldualapp}
\min_{x,\theta}\sup_{u}\left\{u\T[B(Ax-y)-G\T\theta - b]-\frac{1}{2}u\T M u + \mu\one\T\log(c + H\T\theta-C\T u)\right\} + m\log[n_c(\theta)]-\mu\one\T\log(s-S\T\theta)
\end{aligned}
\end{equation}
We introduce dual variable $q$ as a function of the slacks $d$:
\begin{align*}
d &:= \mat{c\\s}-\mat{C\T&-H\T\\0&S\T}\mat{u\\\theta}, \quad D := \text{diagm}(d), 
\quad q := \mu D^{-1}\one,& Q := \text{diagm}(q), \quad 
z := [q,u,x,\theta]\T
\end{align*}
and form the KKT system,
\begin{equation}
%\tag{S.2}
\label{eq:KKT}
F_\mu(z) = \mat{
Dq - \mu\one\\
B(Ax-y) - G\T\theta - b -Mu + \mat{-C&0}q\\
A\T B\T u\\
-Gu + m\nabla\log[n_c(\theta)] + \mat{H&S}q
}
\end{equation}
The Jacobian matrix $F_\mu^{(1)}$ of the system is given by 
\begin{equation}
%\tag{S.3}
\label{eq:KKTJacobian}
\def\arraystretch{1.5}
\nabla F_\mu(z) = \lt[\begin{array}{c|c|c|c}
D & Q\mat{-C\T\\0} &&Q\mat{H\T\\-S\T} \\
\hline
\mat{-C&0}&-M& BA & -G\T\\
\hline
 &A\T B\T  & & \\
\hline
\mat{H&S}& -G & & m\nabla^2\log(n_c)
\end{array}\rt]
\end{equation}
Algorithm \ref{alg:IPsolve} is a damped Newton iteration to find the stationary point of $F_\mu$. At each iteration, we shrink the $\mu$ 
to be a fraction of the current average complementarity conditions, just as in the implementation used in~\cite{JMLR:v14:aravkin13a,aravkin2018generalized}.

We can apply block Gaussian elimination to obtain conditions that make the algorithm implementable. We state the result as a siimple theorem. 

\begin{theo}[IP implementability]
Suppose that the PLQ penalty is nondegenerate, that is $\mbox{null}(C) \cap \mbox{null}(M) =\{0\}$, 
and that the linear model $A$ is full-rank. Then the interior point iteration
\[
p =\nabla F_\mu(z^k)^{-1}F_\mu(z^k)
\]
is implementable when a certain square symmetric system is invertible: 
\begin{equation}
\label{eq:Inv}
\begin{aligned}
T_3 & :=  m\nabla^2\log(n_c) - HQH\T + SQS\T +  ( -G + HD^{-1}QC\T)  T_1^{-1}(-G\T + CD^{-1}QGH\T) \\
 &- ( -G + HD^{-1}QC\T)  T_1^{-1} BA T_2^{-1}A\T  B\T T_1^{-1}(-G\T + CD^{-1}QGH\T)
 \end{aligned}
\end{equation}
\end{theo}
$T_3$ is a symmetric square matrix with dimension equal to the length of the parameter vector $\theta$. 
\begin{proof}
Implementing the IP iteration is equivalent to applying block-Gaussian elimination 
to the system~\eqref{eq:KKTJacobian}. The set of operations is given by 
\[
\begin{aligned}
R_2 &= R_2 + \mat{C&  0} D^{-1}R_1\\
R_4 & = R_4 - \mat{H&  S}  D^{-1} R_1\\
R_3 &= R_3 + A\T  B\T T_1^{-1}R_2\\
R_4 & = R_4 + ( -G + HD^{-1}QC\T)  T_1^{-1} R_2\\
R_4 &= R_4 - ( -G + HD^{-1}QC\T)  T_1^{-1} BA T_2^{-1}R3
\end{aligned}
\]
where we define
\[
\begin{aligned}
T_1 &:= M + CD^{-1}QC\T \\
T_2 & := A\T B\T T_1^{-1}BA
\end{aligned}
\]
The invertibility of $D$ is enforced by Algorithm~\ref{alg:IPsolve}, which keeps slack variables $d$ strictly positive. 
Invertibility of $T_1$ is guaranteed by the nondegeneracy hypothesis for the PLQ, see~\cite[Theorem 14]{JMLR:v14:aravkin13a}. 
Since $B$ must be injective (see Definition~\ref{eq:QSdual}), the full rank condition on $A$ guarantees that 
$T_2$ is invertible. The row operations yield an upper triangular matrix of form 
\begin{equation}
%\tag{S.3}
\label{eq:KKTJacobianC3}
\def\arraystretch{1.5}
\nabla F_\mu(z) = \lt[\begin{array}{c|c|c|c}
D & Q\mat{-C\T\\0} &&Q\mat{H\T\\-S\T} \\
\hline
&-T_1 \T & BA & -G\T + CD^{-1}QGH\T\\
\hline
 & & T_2&   A\T  B\T T_1^{-1}(-G\T + CD^{-1}QGH\T)\\
\hline
& && 
T_3
\end{array}\rt],
\end{equation}
which is invertible if and only if $T_3$ in~\eqref{eq:Inv} is invertible, given the other hypotheses.  
\end{proof}

In the next section, we present synthetic examples and that show the applicability of the approach for inferring simple shape parameters. 

\section{Synthetic Data Experiments}
\label{sec:synthetic}
We illustrate the approach using a {linear} regression example. Consider a regression problem over the quantile Huber family (Figure~\ref{fig:Qhub}) for a data set contaminated by asymmetric errors and outliers. 
The $\tau$ parameter controls slope of the penalty, while $\kappa$ is the robustness threshold. We want to fit the regression model  $x$ as well as obtain the correct parameters $\tau$ and $\kappa$. %Simple residual analysis is not possible {\it a priori}, since the model parameters $x$ are also unknown.

%A self-tuning quantile Huber is flexible penalty, that can effectively adapt to a wide range of errors. In addition, 
{When $\kappa > 0$ in quantile Huber, $\rho(x;\theta)$ is smooth, and we can use the PALM algorithm from Section~\ref{sec:PALM}. 
The quantile Huber penalty is QS, so we can also apply the proposed IP method from Section~\ref{sec:IP}. 
We use both and compare their performance.}
 
The primal form %\footnote{For the dual representation, see caption of Figure~\ref{fig:QuantileQHub}, right panel} 
for the quantile Huber {penalty is}
\begin{equation}
\label{eq:quantileHuberPrimal}
\small
\rho\left(r; \mat{\tau\\\theta}\right) = \begin{cases}
-\tau\kappa r - \frac{(\tau\kappa)^2}{2},& r < -\tau\kappa\\
\frac{1}{2}r^2,& r\in[-\tau\kappa,(1-\tau)\kappa]\\
(1-\tau)\kappa r - \frac{((1-\tau)\kappa)^2}{2},& r > (1-\tau)\kappa.
\end{cases}
\end{equation}
We must choose a parametrization in terms of $\theta$. One option is to take $\theta = [\tau,\kappa]\T$. 
But this parametrization violates assumptions of both first- and second-order approaches in Section~\ref{sec:algo}.
Indeed, $\nabla_\theta \rho(r;\theta)$ would not have a global Lipschitz constant, which is not valid for PALM. 
Neither could we write~\eqref{eq:QSprimaldual} using affine functions of $\theta$. 
Looking carefully at~\eqref{eq:quantileHuberPrimal}, we instead choose 
$\theta_1 = \tau\kappa, \theta_2 = \tau(1-\kappa)$. These new parameters must be non-negative. 
%With any pair $(\theta_1, \theta_2)$ we obtain 
%\[
%\tau = \frac{\theta_1/\theta_2}{1+(\theta_1/\theta_2)}, \quad \kappa = \theta_1  + \theta_2.
%\] 
\begin{table}[h!]
\begin{center}
\setlength{\tabcolsep}{5pt}
\begin{tabular}{c|c|c|c|c}
\hline
$[\tau_t,\kappa_t]$ & $[\tau^*,\kappa^*]$ & $r(x^*)$ & $r(x_{LS})$ & $r(x_M)$ \\ \hline
[0.1,1.0] & {\bf [0.09,1.17]}& {\bf 0.14} & 0.41 &0.26\\ \hline
[0.2,1.0] & {\bf [0.20,1.07]} & {\bf 0.10} & 0.16& 0.13\\ \hline
[0.5,1.0] & {\bf [0.50,0.95]} & {\bf 0.08} & 0.12& 0.09\\ \hline
[0.8,1.0] & {\bf [0.81,1.04]} & {\bf 0.09} & 0.19& 0.11\\ \hline
[0.9,1.0] & {\bf [0.91,1.17]} & {\bf 0.12} & 0.38& 0.36\\
\hline
\end{tabular}
\caption{\small\label{tb:exp} Joint inference of the shape and model parameters for the quantile Huber loss. Columns 2 and 3 show results for proposed method. $r(x) = \|x - x_t\|/\|x_t\|$ denotes relative error. Column 2 contains $\tau, \kappa$ estimated using joint optimization (compare to ground truth simulation parameters in column 1) solved by the proposed IP algorithm with KKT tolerance  set to $10^{-6}$. 
Column 3 shows relative error of the new estimate; compare to Columns 4 and 5, which are relative errors for LS and minimum 1-norm estimates. 
%\red{Why is $\kappa^*$ for the last row so away from 0?}
}
\end{center}
\end{table}

The primal objective can be written as 
\[
\min_{x,\theta\geq 0} \rho(Ax-y;\theta) + m\log[n_c(\theta)],
\]
where $A\in\R^{m\times n}$ {is the design matrix} (always random Gaussian), $x\in\R^n$ is {the model parameter vector}, and $y\in\R^m$ is the observed data vector contaminated by outliers. From Theorem \ref{th:smoothness}, $n_c(\theta)$ is $\mathcal{C}^2$ smooth.
From Theorem \ref{th:convexity}, {the objective in $\theta$ is the sum of a concave term $\rho(Ax-y;\theta)$
and a convex term $m\log[n_c(\theta)]$. The joint problem in $(x,\theta)$ is nonconvex,}
% so we do not even know if the overall objective is biconvex. 
but both first- and second-order methods from Section \ref{sec:algo} can be applied.
\subsection{Shape-Optimized Quantile Huber  vs. Least Square and LAD}
We generate synthetic data with $m = 1000$ and $n = 50$. The measurement errors are sampled from quantile Huber distributions, to {verify} that the approach is able to recover `ground truth' values for $(\tau, \kappa)$ parameters. 
%\begin{itemize}
%\item \texttt{m = 1000}, \texttt{n = 50}
%\item \texttt{A = randn(m,n)}, \texttt{xt = randn(n)}
%\item \texttt{eps}: independent \texttt{m} samples from the distribution defined by quantile-huber penalty.
%\item \texttt{y = A*xt + eps}
%\end{itemize}
We denote ground truth parameters as $x_t$, $\tau_t$, $\kappa_t$, 
while $x^*$, $\tau^*$, $\kappa^*$ are the solutions obtained by solving~\eqref{eq:obj}. 
We provide two reference solutions: $x_{LS}$ is the least square solution, and $x_M$ is 
the solution obtained by solving $\|Ax-b\|_1$.
%these are used as references to evaluate the quality of the final $x$ estimates.
For each $\kappa$ and $\tau$ setting, we run the simulation 10 times, and show the average of the results in Table \ref{tb:exp}. Results shown are obtained by the IP method.

\begin{table}
\begin{center}
\setlength{\tabcolsep}{2pt}
\begin{tabular}{c|c|c|c|c}
\hline
$m$ & $n$ & PALM: Time(s)/\#iter & IP: Time(s)/\#iter & $ f_\text{PALM}^*- f_\text{IP}^*$ \\ \hline
100 &50 & 4.12/96 & 3.11/15 & 7.25e-12\\ \hline
500 & 50 & 5.24/197 & 4.86/16 & 1.38e-08\\ \hline
1000 & 50 & 5.97/112 & 11.68/14 & 1.52e-08\\ \hline
2000 & 100 & 11.41/129 & 69.85/16 & 1.30e-08\\ \hline
2000 & 200 & 22.72/225 & 72.89/16 & 9.61e-08\\ \hline
2000 & 500 & 66.30/394 & 87.85/18 & 3.86e-08\\
\hline
\end{tabular}
\caption{\small\label{tb:runtime} Timing comparison (in seconds) of PALM and IP for the quantile Huber family. Columns 2 and 3 show total run time and number of iterations of PALM and IP; column 3 plots difference in final objective values. IP finds lower objectives, but PALM is faster for larger problems. A large-scale implementation of IP that uses iterative methods in each iteration would help with the scaling issues. 
}
\end{center}
\end{table}
\begin{figure}[h]
\centering
\includegraphics[width=7cm]{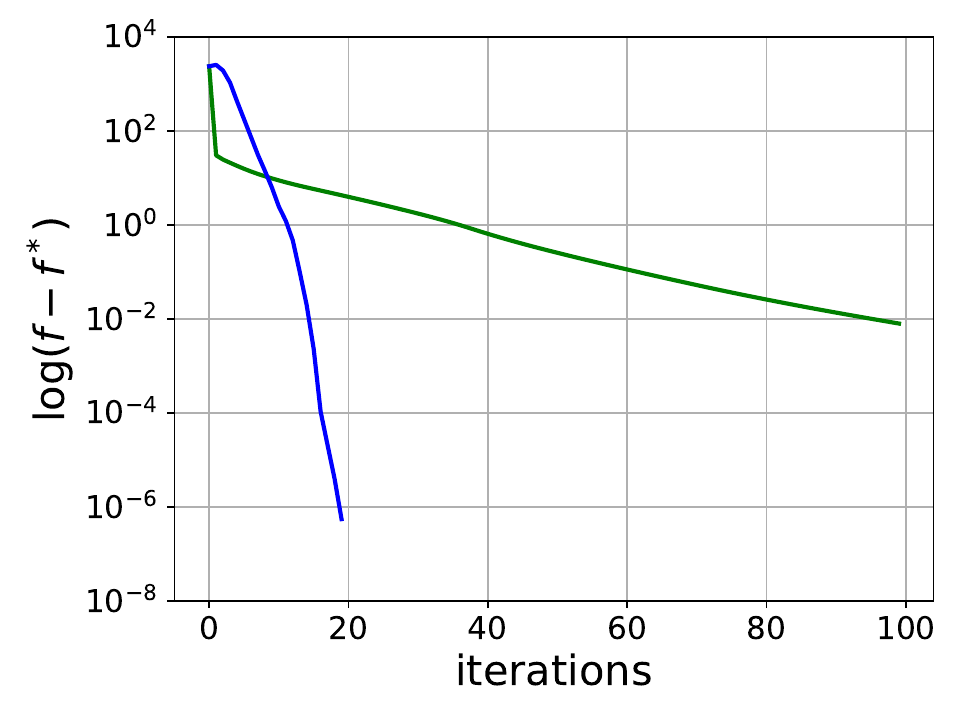}
\caption{\small Convergence history (iterations) for PALM (green) and interior point method (blue). The  
experiment shown is for $\tau = 0.1,\kappa=1$. This behavior is representative; in particular the proposed IP method converges 
in fewer than 20 iterations in all cases. 
% \red{[If possible make the text fonts and labels bigger, add grids, and make lines thicker. Do we need three graphs here? ]}
 }
\label{conhis} 
\end{figure}
The maximum likelihood formulation correctly recovers the shape parameters $(\theta, \tau)$. 
Moreover, the solution $x^*$ obtained from the self-tuned regression is always better 
compared to reference solutions, and the improvement increases as measurement errors become more biased ($\tau$ close 
to $0$ or to $1$). %mpare to least square solution especially when $\tau_t$ is close to $0$ and $1$.
\subsection{PALM vs. IP for Penalty Optimization}
We also compared the performance of PALM and IP over $m \in\{100, 500, 1000, 2000\}$ and 
$n \in \{50, 100, 200, 500\}$, for both iterations and run time. 
The results are shown in Figure~\ref{conhis} and Table~\ref{tb:runtime}. 
From Table~\ref{tb:runtime}, IP converges within 20 iterations independently of problem dimension. However, the total run time exceeds PALM as the problem size increases. This issue can be mitigated by implementing iterative solvers with pre-conditioners; which would be particularly valuable for nonsmooth $\rho$, where PALM cannot be applied.  However, this effort is outside the scope of the paper.

\begin{table}[h!]
\begin{center}
\setlength{\tabcolsep}{3pt}
\begin{tabular}{c|c|c|c|c|c|c}
\hline
$m$ & $n$ & $x$ only: Time (s)/\#iter & Joint: Time(s)/\#iter & $[\kappa^*,\tau^*]$ & Cross-Val Time(s) & $[\kappa_{CV},\tau_{CV}]$\\ \hline
100 &50 & 24.57/4008 & 16.98/2198 & [0.50,1.28] & 585.12 & [0.5,1.5]\\ \hline
500 & 50 & 4.13/430 & 7.31/386 & [0.53,1.03] & 309.96 & [0.5,1.0]\\ \hline
1000 & 50 & 3.78/202 & 7.58/224 & [0.49,1.00] & 349.00 & [0.5,1.0] \\ \hline
2000 & 100 & 14.22/219 & 19.79/255 & [0.49,1.02]& 1026.92 & [0.5,1.0]\\ \hline
2000 & 200 & 29.20/388 & 42.67/513 & [0.49,1.04] & 2315.02 & [0.5,1.0]\\ \hline
2000 & 500 & 158.32/1025 & 202.23/1261 & [0.52,1.20] & 7477.29 & [0.5,1.0]\\
\hline
\end{tabular}
\caption{\small\label{tb:joint_vs_single} Column 3 shows time (s)/iterations for gradient descent on a single instance of the quantile huber problem. Columns 4 and 5 show the time/iterations for PALM for the joint problem over $(x,\theta)$ and resulting estimates of $(\kappa, \tau)$. Columns 6 and 7 shows the time and estimates obtained by a $5 \times 5$ grid search over $(\kappa, \tau)$. Penalty optimization finds good estimates very quickly relative to the baseline. Grid search takes an order of magnitude more time. 
}
\end{center}
\end{table}

\subsection{Penalty Optimization vs. Grid Search}
Finally, we compared the run time and shape parameter estimates for the quantile Huber family with a grid search 
method. We construct a $5\times 5$ grid over $\tau\in[0.1,0.2,0.5,0.8,0.9]$ and $\kappa\in[0.1,0.5,1.0,1.5,1.9]$, split the data into training and validation sets (80\% and 20\%) 
and pick the parameter that performs the best over validation data. 
We also provide the run time for a single instance evaluation with fixed $\theta$ (using gradient descent) as a baseline comparison. 

Results are shown in Table~\ref{tb:joint_vs_single}.
From the table we can see that both methods find good estimates of the shape parameters. However, grid search clearly suffers form the `curse of dimensionality' even at dimensionality 2, as it takes approximately 20 times longer to complete.

%======================================================================
\section{Real Data Example}
\label{sec:real}

In this section, we consider large-scale examples, developing approaches for using `self-tuning' penalties for robust principal component analysis (RPCA). % focusing on PALM- and VP-type algorithms. 
RPCA  has applications to alignment of occluded images \cite{peng2012rasl}, scene triangulation \cite{zhang2012tilt}, model selection \cite{chandrasekaran2009sparse}, face recognition \cite{turk1991face} and document indexing \cite{candes2011robust}.
%\begin{wrapfigure}{r}{0.5\textwidth}
%\centering
%\includegraphics[width=6cm]{./figures/Y1201.eps}
%\caption{\label{bus} figure for background separation}
%\end{wrapfigure}
We develop a self-tuning  background separation approach.
Given a sequence of images (Campus dataset)\footnote{Publicly available at \url{http://vis-www.cs.umass.edu/~narayana/castanza/I2Rdataset/} \label{IR_dataset}},
our goal is to separate the moving objects from the background. % from the moving objects in the video.
We pick $202$ images from the data set, convert them to grey scale and reshape them as column vectors of matrix $Y\in\R^{20480\times202}$.
We model the data $Y$ as the sum of low rank component $L$ and sparse noise $S$; we expect moving objects to be captured by $S$. 

We take advantage of the fact that RPCA is equivalent to regularized Huber regression~\cite{driggs2019adapting}: 
\begin{equation}
\label{eq:RPCAhuber}
\begin{aligned}
\min_{L,S} &\frac{1}{2}\|L+S-Y\|_F^2 + \kappa\|S\|_1 + \lambda\|L\|_* \\
 &= \min_{L} \rho_{\kappa}(Y-L) + \lambda \|L\|_*.
\end{aligned}
\end{equation}

\begin{figure}
\centering
`\begin{tabular}{p{7cm}p{7cm}}
\includegraphics[width=6cm]{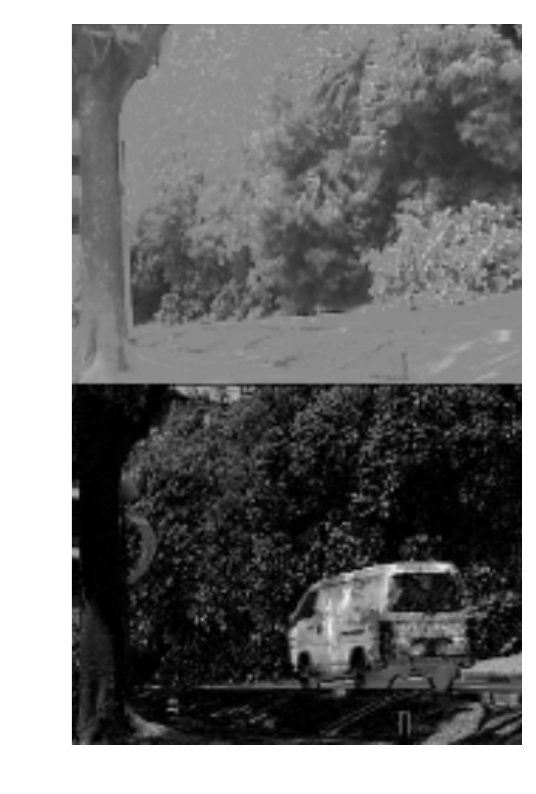}
&
\includegraphics[width=6cm]{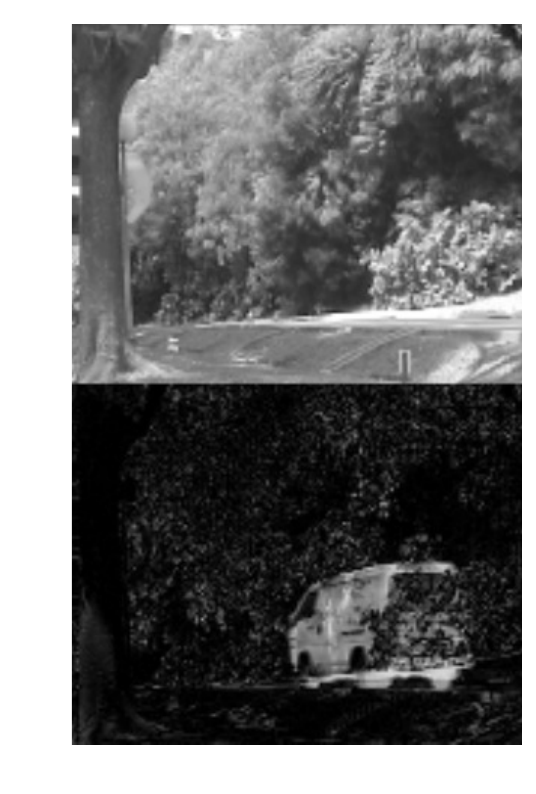}
\\
(a) \small{Huber with $\kappa=0.002,\sigma=1$.}
&(b) \small{Self-tuned Huber, initial: $\kappa=0.002$, $\sigma=1$.}
\end{tabular}
\caption{\small\label{RPCA} RPCA background separation: we optimize over parameters as well as the foreground and background. The separation 
is better as a result of the joint optimization.}
\end{figure}

%
%
%\begin{figure}[h!]
%\centering
%  \begin{tikzpicture}
%  \begin{axis}[
%    thick,
%    height=2cm,
%    width = 3cm,
%    xmin=-0.7,xmax=0.7,ymin=0,ymax=5,
%    no markers,
%    samples=50,
%    axis lines*=left, 
%    axis lines*=middle, 
%    scale only axis,
%    xtick={-0.3,0.3},
%    xticklabels={$-\kappa\sigma$, $\kappa\sigma$},
%    ytick={0},
%    ] 
%\addplot[blue,domain=-0.7:-0.3, very thick]{-6*x+0.5};
%\addplot[blue, domain=-0.3:+0.3, very thick]{ln(1+x^2/0.01)};
%\addplot[red, domain=-0.7:+0.7, dashed]{ln(1+x^2/0.01)};
%\addplot[black, domain=-0.7:-0.05, densely dashed]{-10*x-.25};
%\addplot[black, domain=-0.05:+0.05, densely dashed]{x^2/0.01};
%\addplot[black, domain= .05:0.7, densely dashed]{10*x-.25};
%\addplot[blue,domain=+0.3:+0.7, very thick]{6*x + 0.5};
%\addplot[blue,mark=*,only marks] coordinates {(-0.3,2.3) (0.3,2.3)};
%\end{axis}
%\end{tikzpicture}
%\includegraphics[width=4.7cm]{./figures/dis.eps}
%\caption{\small\label{fig:Tiber} Proposed Tiber (solid blue) interpolates between Student's t (red dash) and 
%Huber (black dash). Empirical CDFS for true residual and fits obtained with 3 penalties; true residual $R = Y - U\T V$ in light blue, 
%Tiber residual in red dashed, $\ell_2$ fit  in blue dashed, and $\ell_1$ fit  in green dashed. Tiber fit matches the 
%empirical residual.
%}
%\end{figure}
%\vspace{-0.1in}
%\end{wrapfigure}
We introduce parameter $\sigma$ {for the Huber penalty} to automatically estimate the right scale of the residual.  The joint $(\kappa, \sigma)$ parametrization is given by 
%Introducing variance parameter is important. Unlike gaussian distribution, in the maximum likelihood formulation, variance parameter won't be just scale in front penalty.
%Instead, it is embedding inside the huber function which is a key factor that controls the shape.
\[
\rho(r;[\kappa,\sigma]) = \begin{cases}
\kappa|r|/\sigma - \kappa^2/2, & |r|>\kappa\sigma\\
r^2/(2\sigma^2), & |r|\le\kappa\sigma.
\end{cases}
\]
We then model the low rank component as $L = U\T V$, where $U\in\R^{k\times m}$ and $V\in\R^{k\times n}$. 
The resulting self-tuning RPCA formulation is given by %(solved by Algorithm~\ref{alg:PALM}):
\[
\min_{U,V,\kappa>0,\sigma>0} \sum_{i,j}\rho(\ip{U_i,V_j} - Y_{i,j};[\kappa,\sigma]) + mn\log[n_c([\kappa,\sigma])].
\]
We solve the problem with PALM, obtaining the result in Figure~\ref{RPCA}(b). As the optimization proceeds,  
$\kappa$ and $\sigma$ decrease to $0$ with a fixed ratio $\alpha = \kappa/\sigma$. 
The self-tuning Huber becomes  the scaled 1-norm,
%\[
%\min_{U,V,S} \|S\|_1\quad \text{s.t. } Y = U\T V + S,
%\]
recovering the original RPCA formulation~\cite{candes2011robust}. The result in Figure~\ref{RPCA}(b) is an improvement 
over the result with initial $(\kappa, \sigma)$ values shown in Figure~\ref{RPCA}(a).
% in particular more background is removed from the sparse component. 

%
%%The weakness of the Huber is that the $\kappa$ has to work well for residuals near the origin as well as in the tail. 
%\subsection{Self-Tuning Tiber}
%The self-tuning approach makes it easy to {design} and automatically tune new penalties. 
%To get additional flexibility, we introduce an inflection point; letting the `slope' near the origin 
% be different from slopes in the tails. We propose the Tiber penalty, shown in Figure~\ref{fig:Tiber}:
%\begin{equation}
%\label{eq:tiber}
%\hspace{-.4cm}
%\begin{cases}
%\frac{2\kappa}{\sigma(\kappa^2+1)}(|r| - \kappa\sigma) + \log(1+\kappa^2), & |r| > \kappa\sigma\\
%\log(1+r^2/\sigma^2),&|r|\le\kappa\sigma
%\end{cases}
%\end{equation}
%%Penalty~\eqref{eq:tiber} is flexible, and can penalize large residuals and small ones d. 
%When we optimize over this parametrized family, the additional flexibility improves the performance relative to the Huber family, see Figure~\ref{RPCA}(d).
%%It is clear the self-tuning approach succeeds, as 
%%the Tiber result at the initial $\kappa, \sigma$ values is useless (Figure~\ref{RPCA}(c)).

\begin{table}[h!]
\begin{center}
\setlength{\tabcolsep}{12pt}
\begin{tabular}{c|c}
\hline
Joint: Time (s)/\#iter & Simple: Time (s)/\#iter \\ \hline
203.11/200 & 126.72/200\\ \hline
210.74/200 & 107.24/200\\
\hline
\end{tabular}
\caption{\small\label{tb:rpca_joint_vs_single} Runtime comparison for self-tuning Huber RPCA (Joint) v.s. single Huber RPCA with fixed $(\kappa, \sigma)$ parameters (Simple). We fix number of iterations to be 200.
}
\end{center}
\end{table}

As in the Section~\ref{sec:synthetic}, we compared the run time of solving the joint (self-tuning) problem with solving a single instance of RPCA. The results are shown in Table~\ref{tb:rpca_joint_vs_single}. In this case, the extended self-tuning problem {\it twice the run time} compared to a single instance. Cross-validation, grid-search and black-box optimization typically require multiple solutions of the original optimization problem to find good parameter values.

\section{Discussion} In this paper we developed a simple approach that extends a regression problem using PLQ penalties to also infer 
unknown shape constraints. We used the statistical interpretation of the PLQ penalty to introduce an additional term
that relates shape constraints to a normalization constant. 

Many existing algorithms can be brought to bear on the extended problem. For smooth penalties, 
the PALM algorithm was quite useful in both synthetic and real examples, particularly for large-scale problems. 
We also developed an interior point method for the PLQ class, that uses conjugate representations of such penalties to 
solve an augmented saddle point system.  

The approach offers several interesting avenues for future research. The nonconvex coupling between convex PLQ penalties and their shape 
parameters is interesting, and finding conditions when the approach is guaranteed to work in general is an open question. 
The maximum likelihood criterion itself is just one approach, and may have limitations as the number of shape parameters grows
with respect to the data. Characterizing these limitations and trying alternatives (such as marginal likelihood) is also left to future work.

%{\bf Remark}
%\begin{itemize}
%\item choose right class of penalty for the application is important
%\item self-tuning shape parameter reduces huge amount of work
%\item from the statistical view, we obtain nice interpretation about the data structure
%\end{itemize}
\bibliographystyle{plain}
\bibliography{nips}

\end{document}